\documentclass[runningheads,a4paper]{llncs}

\usepackage{amssymb}
\usepackage{graphicx}

\usepackage{url}
\newcommand{\keywords}[1]{\par\addvspace\baselineskip
\noindent\keywordname\enspace\ignorespaces#1}

\usepackage{graphicx}
\usepackage{amsmath}
\usepackage{amssymb}
\usepackage{algorithmic}
\usepackage{algorithm}
\usepackage{subfigure}
\usepackage{array}
\usepackage{tabularx}
\usepackage{multirow}

\newcommand{\bfx}{{\textbf{x}}}

\begin{document}

\mainmatter  

\title{Vector Quantization by Minimizing Kullback-Leibler Divergence}

\titlerunning{Vector Quantization by Minimizing Kullback-Leibler Divergence}

\author{Lan Yang$^1$
\thanks{Corresponding author.}
\and
Jingbin Wang$^{2,3}$
\and
Yujin Tu$^4$
\and
Prarthana Mahapatra$^5$
\and
Nelson Cardoso$^6$
}

\authorrunning{Lan Yang et al.}

\institute{$^1$School of Computer Science and Information Engineering,  Chongqing Technology and Business University,
Chongqing 400067, China\\
$^2$Tianjin Key Laboratory of Cognitive Computing and Application, Tianjin
University, Tianjin 300072, China\\
$^3$National Time Service Center, Chinese Academy of Sciences, Xi' an 710600 , China\\
$^4$University at Buffalo, The State University of New York, Buffalo, NY 14260, USA\\
$^5$ Department of Electrical Engineering, M. S. University of Baroda, Vadodara, India \\
$^6$ Instituto Tecnol\'ogico de Aeron¨¢utica, S\"ao Jose dos Campos, SP, Brazil\\
Email: lancy9232001@163.com, jingbinwang1@outlook.com,  yujintu1@yahoo.com}

\maketitle

\begin{abstract}
This paper proposes a new method
for vector quantization
by minimizing the Kullback-Leibler Divergence between the class label distributions over the quantization inputs, which are
 original vectors, and
 the output, which is the quantization subsets of the vector set.
In this way, the vector quantization output can keep as much information
of the class label as possible.
An objective function is constructed and we also developed
an iterative algorithm to minimize it.
The
new method is evaluated on  bag-of-features based image classification problem.
\end{abstract}

\section{Introduction}

Vector quantization is a problem to quantize the
continuous signal vectors into
a discrete dictionary \cite{Pham20092570,Tasdemir20123034,Chang2013233,Zhang20138,Singh2013179,Wang201369}.
The vectors are usually quantized into several discrete quantization subsets,
and then the  index of the subset
will be used as its new representation \cite{Jegou2010316,Jiang2007494}.
It should  compress the original vectors,
and also be helpful for the effective representation of these vectors \cite{Lai20093065}.
An example is the
bag-of-features
based image representation \cite{Lazebnik20062169,Lian20131,Yu2013,Amato2013657,Kashihara2013291,Wang201391}.
In this problem, the local image features are
quantized  to a visual dictionary.
The quantization   is usually conducted in an unsupervised way, assuming that the
class labels are not available \cite{Sprechmann20102042}.
But it's better to learn it by combining the class labels in a supervised way.
The ideal quantization we are seeking is the one that can contains all
the information needed for the classification \cite{Kastner2012256,Yu1990681,AbuMahfouz2005167,wang2012scimate}.

\begin{figure*}[!htb]
\includegraphics[width=\textwidth]{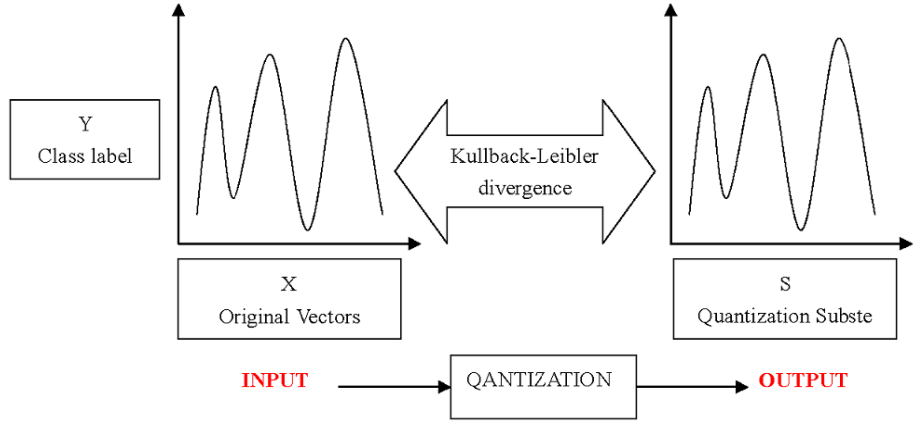}\\
\caption{Minimizing the distribution of class labels over
the original vectors and the quantization subsets.}
\label{Fig1}
\end{figure*}

In this paper, we
present a new
quantization  algorithm so that the class label can be used.
The algorithm is developed by
first modeling the
distribution of the class labels over the quantization input (the original vectors),
and the distribution of the
class labels over the
quantization output (the quantization subsets),
and then minimizing the Kullback-Leibler divergence
between the two distributions \cite{Rached2004917,Dahlhaus1996139,Hershey2007IV317,Liang20131209,Gofman201375,Belov2013141,Park2013,Lee201392,Harmouche2014278}.
We build an objective function to this end, which
involvesthe
quantization of the vectors.
This idea is shown ni Fig. \ref{Fig1}.
Our method is motivated
by Wang et al.'work \cite{wang2013joint}, which learned the
codebook for vector quantization in an supervised way.
However, the differences are of tow-folds:

\begin{itemize}
\item
Wang et al.  \cite{wang2013joint} used
the class label to define
the margin and then maximize it, while
we directly minimize the
Kullback-Leibler divergence between the
class label distributions over the quantization input and output.

\item
Wang et al. \cite{wang2013joint} implement the vector quantization by
using a codebook, while we directly quantize the
vectors into subsets without using a codebook.

\end{itemize}

The rest part of this paper is organized as follows:
Section \ref{sec:met} introduce the
proposed algorithm in details. Section \ref{sec:exp}
shows
the evaluation of the proposed algorithm bag-of-features image classification.
Finally,
Section 6 concludes the paper.

\section{Method}
\label{sec:met}

We first introduce the
distribution of the class label over the
vectors and the quantization subsets in  Section \ref{sec:dist}
and then given the algorithm to minimis the Kullback-Leibler divergence
between these distributions to quantize the vectors in Section \ref{sec:algo}.

\subsection{Class label distributions}
\label{sec:dist}

Assume we have a set of $N$ labeled
training data samples, denoted as
$\{(\bfx_i,y_i)\}_{i=1}^N$.
$\bfx_i\in R^d$ is the $d$-dimensional  vector of the $i$-th sample, and
$y_i\in \mathcal{Y}$ is its corresponding class label.
The classification problem is to predict the class label
of a sample from its vector.
To estimate the class label distribution over the
vector, we define the  $p(y|\bfx)$
as the conditional distribution of class label $y$
over
a sample vector $\bfx$, where $y\in \mathcal{Y}$,
and $\bfx \in \{\bfx_i\}_{i=1}^N$.

\subsubsection{Class label distribution over input}
Given a input  sample $\bfx_i$ and a class label $y$, to estimate $p(y|\bfx_i)$,
we first find its nearest neighbors $\mathcal{N}_i$ \cite{Frentzos2013391,Emrich2013277,Lee2013295,Wang2013136,Taniar20131017},

\begin{equation}
\begin{aligned}
\mathcal{N}_i=k-armin_{\bfx_j} ||\bfx_i-\bfx_j||^2
\end{aligned}
\end{equation}
and then

\begin{equation}
\begin{aligned}
p(y|\bfx_i)=\frac{1}{|\mathcal{N}_i|} \sum_{\bfx_j \in \mathcal{N}_i} I(y_j = y)
\end{aligned}
\end{equation}
where

\begin{equation}
\begin{aligned}
I(y_j = y)
\left\{\begin{matrix}
1 ,& if ~ y_i = y \\
0, & else
\end{matrix}\right.
\end{aligned}
\end{equation}

\subsubsection{Class label distribution over output}
\begin{figure}[!htb]
\centering
  \includegraphics[width=0.6\textwidth]{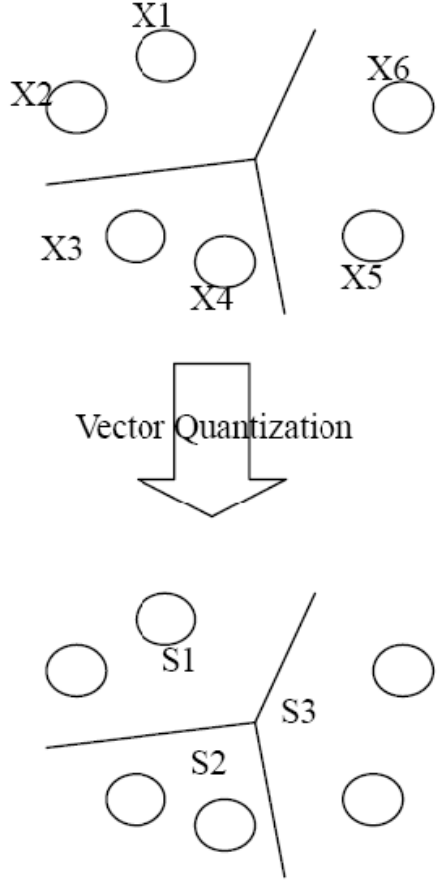}\\
  \caption{Vector quantization from vectors to quantization subsets.}
  \label{Fig2}
\end{figure}

Then we quantize the
vectors  $\{\bfx_i\}_{i=1}^N$ into
$M$ non-overlapping quantization subsets, denoted as
$\{\mathcal{S}_m\}_{m=1}^M$, as shown in Fig. \ref{Fig2}.
Given subset $\mathcal{S}_m$,
and a class label $y\in \mathcal{Y}$,
the
conditional
distribution for $y$ over $\mathcal{S}_m$ is defined as

\begin{equation}
\begin{aligned}
p(y|\mathcal{S}_m)=\frac{1}{|\mathcal{S}_m|} \sum_{\bfx_j \in \mathcal{S}_m} I(y_j = y)
\end{aligned}
\end{equation}

\subsubsection{Kullback---Leibler divergence}
The Kullback---Leibler divergence
between the two distributions $p(y|\mathcal{S}_m)$ and $p(y|\bfx_i)$ are defined as (\ref{equ:1}).

\begin{equation}
\label{equ:1}
\begin{aligned}
D\left (p(y|\bfx)\|p(y|\mathcal{S})\right )
=
\sum_{y\in \mathcal{Y}}
\left \{
\sum_{m=1}^M \left [
\sum_{\bfx_i\in S_m}
p(y|\bfx_i)
log \left (
\frac{p(y|\bfx_i)}{p(y|\mathcal{S}_m)}
\right )
\right ]
\right \}
\end{aligned}
\end{equation}

We try to
have a quantization output $\{\mathcal{S}_m\}_{m=1}^M$ from $\{\bfx_i\}_{i=1}^N$ so that the
the quantization can minimize
the Kullback---Leibler divergence

\begin{equation}
\begin{aligned}
\underset{\mathcal{S}_1,\cdots,S_M}{min}~
\sum_{y\in \mathcal{Y}}
\left \{
\sum_{m=1}^M \left [
\sum_{\bfx_i\in S_m}
p(y|\bfx_i)
log \left (
\frac{p(y|\bfx_i)}{p(y|\mathcal{S}_m)}
\right )
\right ]
\right \}
\end{aligned}
\end{equation}
To solve this problem, we adapted a iterative algorithm.

\subsection{Interactive algorithm}

In this algorithm, we repeat a two-step procedure for many times:
\begin{itemize}
\item In this first step, we fixe the quantization subset as $\mathcal{S}_1^{old},\cdots,S_M^{old}$,
and compute the class label distribution over the quantization output as:

\begin{equation}
\begin{aligned}
p^{new}(y|\mathcal{S}_m)=\frac{1}{|\mathcal{S}_m^{old}|} \sum_{\bfx_j \in \mathcal{S}_m^{old}} I(y_j = y)
\end{aligned}
\end{equation}

\item In the second step, we fixe the distribution and update quantization subset  by (\ref{equ:8}).

\begin{equation}
\label{equ:8}
\begin{aligned}
S_m^{new} =
\left \{
\bfx_i\|
m=\underset{m'=1,\cdots,M}{argmin}
\sum_{y\in \mathcal{Y}}
p(y|\bfx_i)
log \left (
\frac{p(y|\bfx_i)}{p^{new}(y|\mathcal{S}_m)}
\right )
\right \}
\end{aligned}
\end{equation}

\end{itemize}

When a new sample  $\bfx$ not in the training set is given,
it is quantized to the quantization subset as

\begin{equation}
\begin{aligned}
\mathcal{S}_{m^*} \leftarrow \bfx
\end{aligned}
\end{equation}
where

\begin{equation}
\begin{aligned}
m^*=\underset{m=1,\cdots,M}{argmin}
\sum_{y\in \mathcal{Y}}
p(y|\bfx)
log \left (
\frac{p(y|\bfx)}{p^{new}(y|\mathcal{S}_m)}
\right )
\end{aligned}
\end{equation}

\label{sec:algo}

\section{Experiment}
\label{sec:exp}

Experimental evaluation is given in this section on
a real data set with a
bag-of-features based image
classification problem.
We use the Fifteen Natural  Scene Categories database in this experiment \cite{Lazebnik20062169}.
There are 15 classes and for each class, there are around 200 or 300 images.
We use the proposed vector quantization algorithm
to represent the image as a quantization histogram
under the bag-of-features framework.

\begin{figure}[!htb]
\centering
  \includegraphics[width=0.8\textwidth]{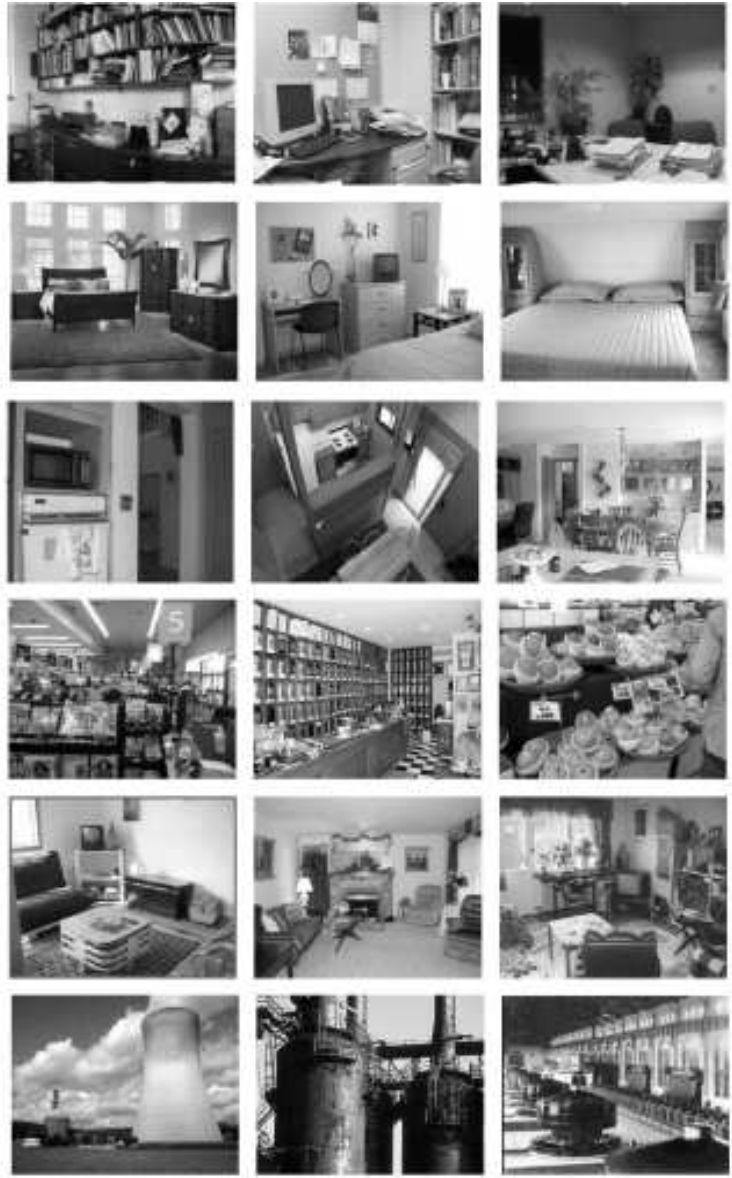}\\
  \caption{Images in the Fifteen Natural  Scene Categories database.}
  \label{Fig3}
\end{figure}

\begin{figure*}[!htb]
\centering
  \includegraphics[width=0.8\textwidth]{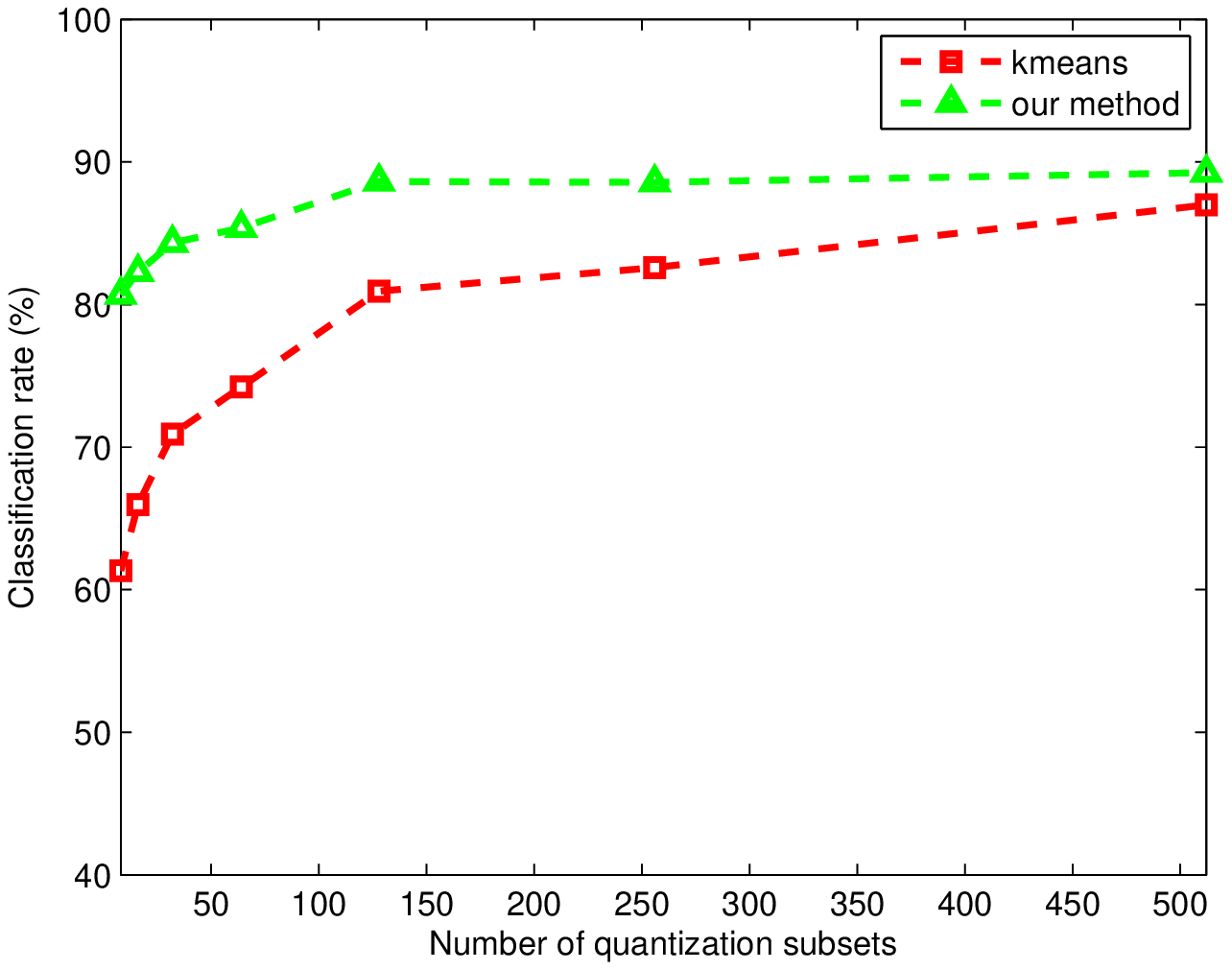}\\
  \caption{Classification results.}
  \label{Fig4}
\end{figure*}

Fig. \ref{Fig4} shows the  results for the Fifteen Natural  Scene Categories data set.
The kmeans algorithm is used as a baseline quantization method \cite{Cuell2009361,Xu2010V26,Barakbah200923,li2011k,zhang20102,wu2009mrec4}.
As in Fig. \ref{Fig4} , the proposed method outperforms the  the
baseline method.

\section{Conclusion}
\label{sec:con}

In this paper, the
problem of vector quantization is investigated.
The idea is motivated by the supervised quantization dictionary learning method
proposed by Wang et al. \cite{wang2013joint}. We try to
keep all information about
the class label from the quantization input to the output.
It is implemented by minimizing Kullback-Leibler Divergence
between the class label distributions   over
quantization input and output. This method can also be applied to social media data analysis \cite{li2014learning,li2013relation,li2009relation}, user recognition of mobile \cite{li2014online}, transportation prediction \cite{li2013real,li2012real}, web news extraction \cite{wu2009web}, malicious websites detection \cite{xu2014evasion,xu2014adaptive,luo2014federated,Zhan20131775,Xu2013141,Xu201230,zhan14characterization}.

\section*{Acknowledgements}

This project was supported by a open research program of the Tianjin Key Laboratory of Cognitive Computing and Application, Tianjin
University, China.


\begin{thebibliography}{10}
\providecommand{\url}[1]{\texttt{#1}}
\providecommand{\urlprefix}{URL }

\bibitem{AbuMahfouz2005167}
Abu-Mahfouz, I.: Drill flank wear estimation using supervised vector
  quantization neural networks. Neural Computing and Applications  14(3),
  167--175 (2005)

\bibitem{Amato2013657}
Amato, G., Falchi, F., Gennaro, C.: On reducing the number of visual words in
  the bag-of-features representation. In: VISAPP 2013 - Proceedings of the
  International Conference on Computer Vision Theory and Applications. vol.~1,
  pp. 657--662 (2013)

\bibitem{Barakbah200923}
Barakbah, A., Kiyoki, Y.: A new approach for image segmentation using
  pillar-kmeans algorithm. World Academy of Science, Engineering and Technology
   59,  23--28 (2009)

\bibitem{Belov2013141}
Belov, D.: Detection of test collusion via kullback-leibler divergence. Journal
  of Educational Measurement  50(2),  141--163 (2013)

\bibitem{Chang2013233}
Chang, C.C., Chou, Y.C., Lin, C.Y.: An indicator elimination method for
  side-match vector quantization. Journal of Information Hiding and Multimedia
  Signal Processing  4(4),  233--249 (2013)

\bibitem{Cuell2009361}
Cuell, C., Bonsal, B.: An assessment of climatological synoptic typing by
  principal component analysis and kmeans clustering. Theoretical and Applied
  Climatology  98(3-4),  361--373 (2009)

\bibitem{Dahlhaus1996139}
Dahlhaus, R.: On the kullback-leibler information divergence of locally
  stationary processes1. Stochastic Processes and their Applications  62(1),
  139--168 (1996)

\bibitem{Emrich2013277}
Emrich, T., Kriegel, H.P., Kr\"oger, P., Niedermayer, J., Renz, M., Z\"ufle,
  A.: Reverse-k-nearest-neighbor join processing. Lecture Notes in Computer
  Science (including subseries Lecture Notes in Artificial Intelligence and
  Lecture Notes in Bioinformatics)  8098 LNCS,  277--294 (2013)

\bibitem{Frentzos2013391}
Frentzos, E., Pelekis, N., Giatrakos, N., Theodoridis, Y.: Cost models for
  nearest neighbor query processing over existentially uncertain spatial data.
  Lecture Notes in Computer Science (including subseries Lecture Notes in
  Artificial Intelligence and Lecture Notes in Bioinformatics)  8098 LNCS,
  391--409 (2013)

\bibitem{Gofman201375}
Gofman, A., Kelbert, M.: Un upper bound for kullback-leibler divergence with a
  small number of outliers. Mathematical Communications  18(1),  75--78 (2013)

\bibitem{Harmouche2014278}
Harmouche, J., Delpha, C., Diallo, D.: Incipient fault detection and diagnosis
  based on kullback-leibler divergence using principal component analysis: Part
  i. Signal Processing  94(1),  278--287 (2014)

\bibitem{Hershey2007IV317}
Hershey, J., Olsen, P.: Approximating the kullback leibler divergence between
  gaussian mixture models. vol.~4, pp. IV317--IV320 (2007)

\bibitem{Jegou2010316}
J\'egou, H., Douze, M., Schmid, C.: Improving bag-of-features for large scale
  image search. International Journal of Computer Vision  87(3),  316--336
  (2010)

\bibitem{Jiang2007494}
Jiang, Y.G., Ngo, C.W., Yang, J.: Towards optimal bag-of-features for object
  categorization and semantic video retrieval. pp. 494--501 (2007), cited By
  (since 1996)99

\bibitem{Kashihara2013291}
Kashihara, K.: Classification of individually pleasant images based on neural
  networks with the bag of features. In: ICOT 2013 - 1st International
  Conference on Orange Technologies. pp. 291--293 (2013)

\bibitem{Kastner2012256}
K\"astner, M., Villmann, T.: Fuzzy supervised self-organizing map for
  semi-supervised vector quantization. Lecture Notes in Computer Science
  (including subseries Lecture Notes in Artificial Intelligence and Lecture
  Notes in Bioinformatics)  7267 LNAI(PART 1),  256--265 (2012)

\bibitem{Lai20093065}
Lai, J., Liaw, Y.C.: A novel encoding algorithm for vector quantization using
  transformed codebook. Pattern Recognition  42(11),  3065--3070 (2009)

\bibitem{Lazebnik20062169}
Lazebnik, S., Schmid, C., Ponce, J.: Beyond bags of features: Spatial pyramid
  matching for recognizing natural scene categories. vol.~2, pp. 2169--2178
  (2006)

\bibitem{Lee201392}
Lee, J., Renard, E., Bernard, G., Dupont, P., Verleysen, M.: Type 1 and 2
  mixtures of kullback-leibler divergences as cost functions in dimensionality
  reduction based on similarity preservation. Neurocomputing  112,  92--108
  (2013)

\bibitem{Lee2013295}
Lee, K.W., Choi, D.W., Chung, C.W.: Dart: An efficient method for
  direction-aware bichromatic reverse k nearest neighbor queries. Lecture Notes
  in Computer Science (including subseries Lecture Notes in Artificial
  Intelligence and Lecture Notes in Bioinformatics)  8098 LNCS,  295--311
  (2013)

\bibitem{li2014learning}
Li, H., Cao, T., Li, Z.: Learning the information diffusion probabilities by
  using variance regularized em algorithm. In: IEEE/ACM International
  Conference on Advances in Social Networks Analysis and Mining. pp. 273--280
  (2014)

\bibitem{li2012real}
Li, H., Li, Z., White, R., Wu, X.: A real-time transportation prediction
  system. In: The 25th International Conference on Industrial, Engineering \&.
  Other Applications of Applied Intelligent Systems, vol. 7345, pp. 68--77.
  Springer Berlin Heidelberg (2012)

\bibitem{li2013real}
Li, H., Li, Z., White, R.T., Wu, X.: A real-time transportation prediction
  system. Applied intelligence  39(4),  793--804 (2013)

\bibitem{li2009relation}
Li, H., Wu, G.Q., Hu, X.G., Wu, X., Bi, Y.J., Li, P.P.: A relation extraction
  method of chinese named entities based on location and semantic features. In:
  IEEE International Conference on Granular Computing. pp. 334--339. IEEE
  (2009)

\bibitem{li2011k}
Li, H., Wu, G.Q., Hu, X.G., Zhang, J., Li, L., Wu, X.: K-means clustering with
  bagging and mapreduce. In: The 44th Hawaii International Conference on System
  Sciences. pp. 1--8. IEEE (2011)

\bibitem{li2014online}
Li, H., Wu, X., Li, Z.: Online learning with mobile sensor data for user
  recognition. In: The 29th Symposium On Applied Computing. pp. 64--70. ACM
  (2014)

\bibitem{li2013relation}
Li, H., Wu, X., Li, Z., Wu, G.: A relation extraction method of chinese named
  entities based on location and semantic features. Applied Intelligence
  38(1),  1--15 (2013)

\bibitem{Lian20131}
Lian, Z., Godil, A., Sun, X., Xiao, J.: Cm-bof: visual similarity-based 3d
  shape retrieval using clock matching and bag-of-features. Machine Vision and
  Applications pp. 1--20 (2013)

\bibitem{Liang20131209}
Liang, Z., Li, Y., Xia, S.: Adaptive weighted learning for linear regression
  problems via kullback-leibler divergence. Pattern Recognition  46(4),
  1209--1219 (2013)

\bibitem{luo2014federated}
Luo, W., Xu, L., Zhan, Z., Zheng, Q., Xu, S.: Federated cloud security
  architecture for secure and agile clouds. In: High Performance Cloud Auditing
  and Applications, pp. 169--188. Springer (2014)

\bibitem{Park2013}
Park, S., Shin, M.: Kullback-leibler information of a censored variable and its
  applications. Statistics  (2013)

\bibitem{Pham20092570}
Pham, T., Brandl, M., Beck, D.: Fuzzy declustering-based vector quantization.
  Pattern Recognition  42(11),  2570--2577 (2009)

\bibitem{Rached2004917}
Rached, Z., Alajaji, F., Campbell, L.: The kullback-leibler divergence rate
  between markov sources. IEEE Transactions on Information Theory  50(5),
  917--921 (2004)

\bibitem{Singh2013179}
Singh, A., Murthy, K.: Neuro-curvelet model for efficient image compression
  using vector quantization. Lecture Notes in Electrical Engineering  258 LNEE,
   179--185 (2013)

\bibitem{Sprechmann20102042}
Sprechmann, P., Sapiro, G.: Dictionary learning and sparse coding for
  unsupervised clustering. pp. 2042--2045 (2010)

\bibitem{Taniar20131017}
Taniar, D., Rahayu, W.: A taxonomy for nearest neighbour queries in spatial
  databases. Journal of Computer and System Sciences  79(7),  1017--1039 (2013)

\bibitem{Tasdemir20123034}
Tasdemir, K.: Vector quantization based approximate spectral clustering of
  large datasets. Pattern Recognition  45(8),  3034--3044 (2012)

\bibitem{Wang201391}
Wang, C., Zhang, B., Qin, Z., Xiong, J.: Spatial weighting for bag-of-features
  based image retrieval. Lecture Notes in Computer Science (including subseries
  Lecture Notes in Artificial Intelligence and Lecture Notes in Bioinformatics)
   8032 LNAI,  91--100 (2013)

\bibitem{wang2013joint}
Wang, J.J.Y., Bensmail, H., Gao, X.: Joint learning and weighting of visual
  vocabulary for bag-of-feature based tissue classification. Pattern
  Recognition  46(12),  3249--3255 (2013)

\bibitem{Wang2013136}
Wang, J.S., Lin, C.W., Yang, Y.: A k-nearest-neighbor classifier with heart
  rate variability feature-based transformation algorithm for driving stress
  recognition. Neurocomputing  116,  136--143 (2013)

\bibitem{Wang201369}
Wang, W.J., Huang, C.T., Liu, C.M., Su, P.C., Wang, S.J.: Data embedding for
  vector quantization image processing on the basis of adjoining state-codebook
  mapping. Information Sciences  246,  69--82 (2013)

\bibitem{wang2012scimate}
Wang, Y., Jiang, W., Agrawal, G.: {SciMATE: A Novel MapReduce-Like Framework
  for Multiple Scientific Data Formats}. In: Cluster, Cloud and Grid Computing
  (CCGrid), 2012 12th IEEE/ACM International Symposium on. pp. 443--450. IEEE
  (2012)

\bibitem{wu2009web}
Wu, G.Q., Wu, X., Hu, X.G., Li, H., Liu, Y., Xu, R.G.: Web news extraction
  based on path pattern mining. In: The Sixth International Conference on Fuzzy
  Systems and Knowledge Discovery. vol.~7, pp. 612--617. IEEE (2009)

\bibitem{wu2009mrec4}
Wu, G., Li, H., Hu, X., Bi, Y., Zhang, J., Wu, X.: Mrec4. 5: C4. 5 ensemble
  classification with mapreduce. In: The Fourth ChinaGrid Annual Conference.
  pp. 249--255. IEEE (2009)

\bibitem{Xu2010V26}
Xu, J., Liu, H.: Web user clustering analysis based on kmeans algorithm.
  vol.~2, pp. V26--V29 (2010)

\bibitem{Xu2013141}
Xu, L., Zhan, Z., Xu, S., Ye, K.: Cross-layer detection of malicious websites.
  In: CODASPY 2013 - Proceedings of the 3rd ACM Conference on Data and
  Application Security and Privacy. pp. 141--152 (2013)

\bibitem{xu2014evasion}
Xu, L., Zhan, Z., Xu, S., Ye, K.: An evasion and counter-evasion study in
  malicious websites detection. In: Communications and Network Security (CNS),
  2014 IEEE Conference on. pp. 265--273. IEEE (2014)

\bibitem{Xu201230}
Xu, S., Lu, W., Zhan, Z.: A stochastic model of multivirus dynamics. IEEE
  Transactions on Dependable and Secure Computing  9(1),  30--45 (2012)

\bibitem{xu2014adaptive}
Xu, S., Lu, W., Xu, L., Zhan, Z.: Adaptive epidemic dynamics in networks:
  Thresholds and control. ACM Transactions on Autonomous and Adaptive Systems
  (TAAS)  8(4), ~19 (2014)

\bibitem{Yu1990681}
Yu, G., Russell, W., Schwartz, R., Makhoul, J.: Discriminant analysis and
  supervised vector quantization for continuous speech recognition. vol.~2, pp.
  681--688 (1990)

\bibitem{Yu2013}
Yu, J., Qin, Z., Wan, T., Zhang, X.: Feature integration analysis of
  bag-of-features model for image retrieval. Neurocomputing  (2013)

\bibitem{Zhan20131775}
Zhan, Z., Xu, M., Xu, S.: Characterizing honeypot-captured cyber attacks:
  Statistical framework and case study. IEEE Transactions on Information
  Forensics and Security  8(11),  1775--1789 (2013)

\bibitem{zhan14characterization}
Zhan, Z., Xu, M., Xu, S.: A characterization of cybersecurity posture from
  network telescope data. In: Proceedings of The 6th International Conference
  on Trustworthy Systems, InTrust. vol.~14

\bibitem{Zhang20138}
Zhang, B., Zhou, Y., Pan, H.: Vehicle classification with confidence by
  classified vector quantization. IEEE Intelligent Transportation Systems
  Magazine  5(3),  8--20 (2013)

\bibitem{zhang20102}
Zhang, J., Wu, G., Li, H., Hu, X., Wu, X.: A 2-tier clustering algorithm with
  mapreduce. In: The Fifth Annual ChinaGrid Conference. pp. 160--166. IEEE
  (2010)

\end{thebibliography}
\end{document}